\documentclass[10pt]{article}
\usepackage{microtype}
\usepackage{graphicx}
\usepackage{subfigure}
\usepackage{booktabs}
\usepackage{mdframed}
\usepackage{hyperref}

\usepackage{iclr2025_conference,times}
\usepackage{amsmath}
\usepackage{amssymb}
\usepackage{mathtools}
\usepackage{amsthm}

\usepackage[capitalize,noabbrev]{cleveref}

\theoremstyle{plain}

\theoremstyle{definition}

\theoremstyle{remark}

\usepackage[disable,textsize=tiny]{todonotes}
\author{Oleksandr Balabanov  \\
Stockholm University \\
Department of Physics \\
Stockholm, Sweden \\
\texttt{oleksandr.balabanov@fysik.su.se} \\
\And
Hampus Linander \\
Department of Mathematical Sciences \\
Chalmers university of technology \& University of Gothenburg \\
Gothenburg, Sweden \& \\
VERSES Research Lab \\
Los Angeles, California, USA \\
\texttt{hampus.linander@chalmers.se}
}
\title{Uncertainty quantification in fine-tuned LLMs using LoRA ensembles}

\iclrfinalcopy

\begin{document}
\maketitle

\begin{abstract}Fine-tuning large language models can improve task specific performance, although a general understanding of what the fine-tuned model has learned, forgotten and how to trust its predictions is still missing. We derive principled uncertainty quantification for fine-tuned LLMs with posterior approximations using computationally efficient low-rank adaptation ensembles. We analyze three common multiple-choice datasets using low-rank adaptation ensembles based on Mistral-7b, and draw quantitative and qualitative conclusions on their perceived complexity and balance between retained prior knowledge and domain specific adaptation during and after fine-tuning. We identify unexpected retention of acquired knowledge during fine-tuning in the overfitting regime. Code: {\it github.com/oleksandr-balabanov/equivariant-posteriors}
\end{abstract}

\section{Introduction} \label{Introduction}
Large language models (LLMs) learns conditional distributions of vocabularies useful for generative tasks, text classification and code completion. Trained on a corpus of sequences such as natural language text and program source code, these models have shown remarkable capabilities \citep{bubeck2023sparks,touvron2023llama,touvron2023llama2,jiang2023mistral}. As the applications of LLMs continue to explode, informed use of these models hinge on a proper understanding of the uncertainty of their output \citep{huang_look_2023,kuhn_semantic_2023,malinin_uncertainty_2021,ren_out--distribution_2023}.

To align an LLM towards particular needs, such as answering factual questions and instructions, a common approach is to fine-tune the model using a specialised dataset targetting human interaction \citep{ouyang2022training}. Fine-tuning involves additional training of a pre-trained LLM on a smaller dataset generated with humans \citep{zhong2021adapting} or by other LLMs \citep{peng2023instruction,zhang2023llama}.  By fine-tuning, the model adapts its parameters to better capture the nuances, vocabulary, and style of the target domain\citep{peng2023instruction}. 

Even full-model fine-tuning often requires orders of magnitude less training time, but still incur the same computational complexity and memory usage. Prior studies have shown that pre-trained language models can effectively adapt to specific tasks within smaller parameter subspaces \cite{li2018measuring, aghajanyan2020intrinsic}, indicating an inherently low rank during the fine-tuning process.
To further reduce the training time and computational complexity of fine-tuning, a common method used is Low-Rank Adaptation (LoRA) \citep{hu2021lora}. As one member of the family of methods typically referred to as parameter efficient fine-tuning (PEFT), LoRA effectively reduces the number of parameters requiring training by keeping the pre-trained weights unchanged and integrating low-rank trainable matrices into each layer of the LLM transformer architecture.  

Key questions arise after fine-tuning: What areas of knowledge remain outside the model's expertise? What knowledge is retained from the pre-trained model? What knowledge is gained during the fine-tuning process on the target dataset?  These questions are fundamental in  guiding us towards a more reliable, interpretable and trustworthy application of LLMs. 

We contribute to addressing these questions by analyzing the uncertainty in LLMs that have been fine-tuned using LoRA for answering multiple-choice question answers (QAs). These tasks involve definite single-token target labels, which significantly simplifies the analysis while still requiring the model to have a thorough understanding of context. We use predictive entropy and mutual information to quantify inherent- and model uncertainty. These entropic uncertainty measures are calculated for a Bayesian posterior derived from an ensemble of LLMs fine-tuned with LoRA.

Using the pre-trained Mistral-7b \cite{jiang2023mistral} model as a prior, we fine-tune  on the CommonsenseQA (CQA) \cite{talmor-etal-2019-commonsenseqa} and evaluate its performance on both CQA and the Social Sciences and STEM components of the MMLU \cite{hendryckstest2021} multiple-choice QA datasets. By quantifying the evolution of knowledge using predictive entropy and mutual information, we demonstrate how these measures can be used to reason about the balance between retained prior knowledge and acquired domain specific knowledge during and after fine-tuning.
\section{Contributions} \label{Our contribution}

\begin{itemize}
    \item We show how uncertainty quantification with entropic measures can be used to reason about the balance between retained knowledge and domain specific adaptation.
    
    \item We derive posterior approximations for LLMs fine-tuned on target datasets using ensembles of LoRA  members. On the way, we provide a Bayesian interpretation of fine-tuning, early-stopping and conditional generative tasks. 
    
    \item We analyze three common multi-choice QA datasets, CQA, MMLU STEM and MMLU Social Sciences, using LoRA ensemble posteriors derived from a pre-trained Mistral-7b model, and identify unexpected retention of acquired knowledge in the overfitting regime. 
\end{itemize}

\section{Related Work} \label{Related Work}

\subsection{Uncertainty quantification in LLMs} 

There has been significant interest in uncertainty quantification across various tasks and domains in neural networks~\cite{Gal2015, Gal2016, malinin2018predictive, ovadia2019trust, malinin2021uncertainty, 2022arXiv220514334L, kuhn_semantic_2023, 2023arXiv230519187L}. 

This interest extends to the realm of Large Language Models (LLMs), where accurately quantifying prediction uncertainty is a key focus \cite{xiao-etal-2022-uncertainty, 2022arXiv220514334L, mielke-etal-2022-reducing, chen2023quantifying, duan2023shifting, huang_look_2023, shorinwa2024surveyuncertaintyquantificationlarge}. The application of LLMs in generative tasks introduces unique challenges, notably in measuring uncertainty of the generative outputs.~\cite{liu2019roberta, malinin_uncertainty_2021, kuhn_semantic_2023, 2023arXiv230519187L}. The disentanglement of uncertainty into aleatoric and epistemic was recently discussed in the context of LLMs~\cite{hou2023decomposing,ling2024uncertainty}. However, it was done via ensembling of the model inputs rather than the model instances, and not in the context of fine-tuning tasks.  

\subsection{Fine-tuning in LLMs} 
 Fine-tuning has recently become an integral part of the LLM ecosystem where it is used to target specific tasks such as general instruction-following model, through methods like Reinforcement Learning from Human Feedback~\cite{pmlr-v97-houlsby19a, hu2021lora, liu2019roberta, ding2022delta, Ding2023}. 

The large computational demands of training and fine-tuning LLMs have  resulted in development of more efficient techniques, known as parameter-efficient fine-tuning (PEFT)\cite{liu2022fewshot, ding2022delta, Ding2023, shi2024dept}. These methods typically involve training a small number of additional parameters on top of a fixed, pre-trained LLM, with a prominent approach being the use of low-rank adapters (LoRA) for each weight matrix~\cite{hu2021lora}.

\subsection{LLMs within Bayesian methods}

Bayesian inference techniques in neural networks have a long-standing history \cite{Denker1987, Tishby1989, Buntine1991, MacKay1991}. These methods establish a systematic approach to derive reliable and interpretable uncertainty estimates. Previous research on Bayesian language models has been primarily concentrated on the pretraining phase of language models, rather than on their fine-tuning \cite{NEURIPS2019_154ff894, inproceedingsXue, cinquin2021pathologies, zhang2018mixup, chen2024calibrating}.

Recent studies have begun exploring the fine-tuning of language models using a Bayesian approach. For instance, in \cite{NEURIPS2020_bcff3f63} and \cite{zhang2021bayesian}, the attention modules are sampled either from simplex-constrained attention distributions or using Bayesian Belief Networks. Neither of these studies address entropic uncertainty measures, nor do they utilize ensembling or PEFT methods for posterior approximation. 

\cite{yang2024bayesian} employs a post-hoc Laplace approximation \cite{MacKay1992} to model LoRA parameters for fine-tuning. While this study does use LoRA for fine-tuning, it does not explore entropic uncertainty measures and focuses on the posterior over LoRA parameters rather than the model as a whole, which could limit straightforward Bayesian interpretability. There are also strong indications that deep ensembles provide more accurate posteriors compared to single-model stochastic methods like Laplace and Monte Carlo dropout \cite{Gustafsson2019, Ovadia2019, Lakshminarayanan2019, Wilson2020, Dwaracherla2022, Balabanov_2023_CVPR}. For a more detailed discussion, see Sec.~\ref{Lora_posterior_quality}.

\subsection{Ensembling LLMs}

Two recent studies have explored the use of ensembling in fine-tuning LLMs, specifically focusing on full model fine-tuning method where all weights are optimized \cite{gleave2022uncertainty, sun2022quantifying}. This approach has large memory overhead by construction. While uncertainty quantification using variance across the ensembles is considered, their methods lack a Bayesian formalism that could provide a more grounded interpretation and understanding.

An alternative method, BatchEnsemble \cite{wen2020batchensemble}, employs a base model with modifications through multiplicative, component-specific rank-1 matrices. This method has been applied to LLMs, but in the context of pre-training rather than fine-tuning \cite{tran2022plex}.

There are also recent endeavors employing LoRA ensembles for fine-tuning LLMs \cite{wang2023lora, zhai2023uncertaintypenalized}. Although these studies address uncertainty quantification, they do so without adopting a Bayesian framework and do not clearly distinguish between epistemic and aleatoric components, which is critical for data interpretability. In another work \cite{arteaga2024hallucinationdetectionllmsfast}, a separate predictor is trained on the LoRA ensembles uncertainty metrics to detect hallucinations. \cite{niu2024functionalleveluncertaintyquantificationcalibrated} introduces a mixture-of-expert framework (UQ4CT) to explicitly capture and calibrate model uncertainty during fine-tuning. All these works are highly relevant yet differ from ours, which provides a principled Bayesian treatment for LoRA ensembles during fine-tuning and examines the evolution of uncertainty metrics to draw conclusions about the balance between retained prior knowledge and newly acquired domain adaptation.

\section{Bayesian Deep Learning for LLMs} 

When formulated as a conditional generative model, an LLM can be viewed as a function taking a sequence of tokens \( s^* \) as input, and predicting a distribution over the set of possible tokens \( t^* \) from the vocabulary as output. Such a model is typically trained to predict the next token in a sequence. 

The prediction should be consistent with our observed data. This is encapsulated in the predictive probability distribution \( p(t^*| s^*, \mathcal{D}) \), with training data \( \mathcal{D} = \{s_i, t_i\}_{i=1}^N \), giving a conditional distribution over tokens \( t^* \). To obtain a specific prediction, we can sample from this distribution, for example, using maximum posterior estimation, \( t^* = \arg \max_{t} p(t| s^*, \mathcal{D}) \). 

Given a model with parameters \(\theta\), the predictive distribution
\begin{equation}
p\bigl(t^* \mid s^*, \mathcal{D}\bigr) 
= \int p\bigl(t^* \mid s^*, \theta\bigr)\, p\bigl(\theta \mid \mathcal{D}\bigr)\, d\theta
\end{equation}
is obtained by marginalizing over the posterior distribution of the parameters. 
The posterior \(p(\theta \mid \mathcal{D})\) represents the uncertainty associated with model parameters \( \theta \), quantifying how likely each choice of \(\theta\) is, given the observed data. 
For a fixed \(\theta\), the term \(p(t^* \mid s^*, \theta)\) captures data uncertainty, reflecting the variability in the model's output \(t^*\) for the input \(s^*\). 
By integrating over all possible parameter values \(\theta\), the final predictive distribution combines both uncertainty in the parameters with the inherent variability in the data~\cite{Gawlikowski2023}.

\subsection{Fine-tuning} 
The objective of fine-tuning is to use a specialized dataset to further improve the knowledge of a generally pre-trained model for a given task. This process can be conceptualized as a conditional generative task on a target domain, denoted as $\mathcal{D}_\text{fine-tuning}$. 

Fine-tuning assumes prior knowledge about the models parameters, expressed as $p(\theta)$. This prior is typically centered around the pre-trained models parameters, $\omega_\text{pre-trained}$, to retain the existing knowledge and apply it to the target domain. We assume a normal prior distribution centered on the pre-trained parameter values: $p(\theta)=N(\theta; \omega_\text{pretrained}, \lambda^{-1}I_{\text{dim}[\theta]})$ with diagonal covariance $I_{\text{dim}[\theta]}$. By adjusting the prior variance $\lambda^{-1}$, we can control the extent to which the model is allowed to deviate from the pre-trained parameter values. In practice, this involves finding a balance between two extremes: a small $\lambda^{-1}$ means that the fine-tuning is largely disregarded, whereas a large $\lambda^{-1}$ implies that the pre-trained models knowledge is not utilized. As with any posterior considerations in a Bayesian formalism, fine-tuning requires careful adjustment of this parameter to target optimal performance for the target tasks.

Note that in this formulation, there is no emphasis on the origin of the pre-trained parameters. Typically, for large language models, the training procedure consists of maximum likelihood optimization combined with supervised reinforcement learning \cite{touvron2023llama2} and the specific details of the training procedure are often not publicly disclosed. In our formulation, we do not need to delve into these origins and instead formally take it as prior knowledge upon which we base our fine-tuning.

\subsection{Posterior approximation methods}
Variational inference \cite{Barber1998, Graves2011, Blundell2015} offers a computationally efficient approximation method for the Bayesian posterior. It involves exploring a family of distributions \( q_\omega(\theta) \), parameterized by \( \omega \), to approximate the posterior \( p(\theta|\mathcal{D}) \). A prevalent approach to identify the optimal \( q_\omega(\theta) \) in this set involves minimizing the Kullback-Leibler (KL) divergence \( \text{KL}(q_\omega(\theta) \,||\, p(\theta|\mathcal{D})) \) \cite{Gawlikowski2023}. This divergence, which measures information loss when \( q_\omega(\theta) \) approximates \( p(\theta|\mathcal{D}) \), can be expressed as:

\begin{align} 
\begin{split} 
&\text{KL}(q_\omega(\theta)\,||\,p(\theta|\mathcal{D}) ) =\int \, d\theta \,  q_\omega(\theta) \, \log\,\frac{q_\omega(\theta)}{p(\theta|\mathcal{D})}  \\
&= \text{KL}(q_\omega(\theta)\,||\,p(\theta) ) - \mathbb{E}_{q_\omega(\theta)} [\log \, p(t | \theta, s)] + C,\\
\end{split}
\label{eq:VI}
\end{align}
where \( \mathcal{D} = \{(t, s)\} \) represents the training data, and \( C \) is a constant term independent of \( \theta \). Here we applied Bayes' theorem $p(\theta|t,s)  = p(t |\theta, s) \, p(\theta) / p(t |s)$. Minimizing the loss function given in Eq.~(\ref{eq:VI}) results in an approximation of the posterior \( p(\theta|\mathcal{D}) \). This loss function includes two components: (1) the KL divergence against the prior \( p(\theta) \) and (2) the expected negative log likelihood (ENLL), a common loss metric for classification tasks. 

The optimization problem described in Eq.~(\ref{eq:VI}) does not specify the origin of the in-domain (observed) data $\mathcal{D}$, and it remains equally valid for a fine-tuning task when substituting $\mathcal{D} = \mathcal{D}_\text{fine-tune}$. However, it's important to stress that for a fine-tuning task, the prior is assumed to be centered around the pre-trained solution $\omega_\text{pretrained}$, whereas for a typical training procedure, the prior is usually considered to be centered around the origin in parameter space.

\subsection{Early Stopping}

Early stopping can be used to enhance generalization to unseen in-domain data, thereby reducing the effect of overfitting. 

In the context of Bayesian deep learning, the objective is to approximate the posterior distribution \( p(\theta|\mathcal{D}) \), where \( \mathcal{D} \) represents the entire task domain. Access to the full domain \( \mathcal{D} \) is typically not available, and we work with a subset \( \mathcal{D}^{\text{train}} \). Generally, \( p(\theta|\mathcal{D}^{\text{train}}) \) does not closely mirror \( p(\theta|\mathcal{D}) \), leading to overfitting, where \( p(\theta|\mathcal{D}^{\text{train}}) \) only acknowledges the training data and struggles to generalize.

Early stopping improves the approximation of \( p(\theta|\mathcal{D}) \). Within the variational inference picture, \( q_\omega(\theta) \) is optimized by minimizing \( \text{KL}(q_\omega(\theta)\,||\,p(\theta|\mathcal{D}^\text{train}) ) \), but evaluated on the entire task domain \( \mathcal{D} \) using \( \text{KL}(q_\omega(\theta)\,||\,p(\theta|\mathcal{D}) ) \).  The prior contribution to the KL loss \( \text{KL}(q_\omega(\theta)\,||\,p(\theta|\mathcal{D}) ) \) can be dropped assuming the entire task domain \( \mathcal{D} \) to have a large number of samples. In practice, the ENLL contribution to the KL loss is estimated using the validation dataset \( \mathcal{D}^{\text{val}} \),  assuming it represents the entire task domain effectively and was not part of the training. 

\section{Deep Ensembles with Low-Rank Adaptation} \label{submission}

\subsection{Low-Rank Adaptation} 

Low-Rank Adaptation (LoRA) provides a computationally efficient alternative to full-model fine-tuning by keeping the pre-trained weight matrices static and adding trainable low-rank matrices into each transformer layer, thereby decreasing the total number of parameters requiring training \cite{hu2021lora}.

Given a pre-trained weight matrix \( W_{\text{pretrained}} \) in the space \( \mathbb{R}^{d \times k} \), the adaptations are represented as \( W_{\text{pretrained}} + BA \), where \( B \) resides in \( \mathbb{R}^{d \times r} \), \( A \) in \( \mathbb{R}^{r \times k} \), and the rank \( r \) is substantially smaller than both \( d \) and \( k \). Throughout this training approach, the variables \( A \) and \( B \) are tuned, while \( W_{\text{pretrained}} \) is kept unchanged.

\subsection{LoRA Deep Ensembles}

We use LoRA ensembles to approximate the posterior \( p(\theta | \mathcal{D}_{\text{fine-tune}}) \). Ensembles of LoRA members have been introduced recently \cite{wang2023lora}, here we provide a Bayesian treatment.

Deep ensembles can be explicitly interpreted in a Bayesian variational framework by assuming an ansatz consisting of a mixture of sharply peaked Gaussians centered at each parameter realization \(\omega_k\), with \(k\) indexing ensemble members \cite{Hoffmann2021, Balabanov_2023_CVPR}. With this ansatz, the variational objective reduces to the sum of each member’s log-likelihood term plus an L2 regularization. 

For additional context, deep ensembles can alternatively be viewed as a particle-based variational inference scheme, where each member evolves under Wasserstein gradient descent to minimize the KL divergence \cite{Angelo2021, wild2023rigorouslinkdeepensembles}. Both interpretations naturally give rise to a repulsive contribution, stemming from the prior term in the KL divergence, that encourages diversity and improves the posterior approximation \cite{Balabanov_2023_CVPR, Angelo2021, wild2023rigorouslinkdeepensembles}. In the present work, we omit this repulsive term and defer its exploration to future research. 

We build on the variational deep‐ensemble framework of Hoffmann et al.\cite{Hoffmann2021} and Balabanov et al.\cite{Balabanov_2023_CVPR}, extending it to incorporate the LoRA reparametrization. The trainable low-rank LoRA matrices \( A_k \) and \( B_k \) are not treated as a subset of the variational parameters \( \omega_k \) of the deep ensemble ansatz. Rather, they are auxiliary degrees of freedom that parameterize these variational parameters. That is, we multiply matrices \( A_k \) and \( B_k \) to obtain \( W_k =  (W_\text{pretrained} + A_k B_k) \subset \omega_k \), and \( W_k \) is used in the deep ensemble ansatz for approximating the posterior distribution. The LoRA parametrisation helps us to efficiently and inexpensively find good values for \( W_k \subset \omega_k 
 \).

\subsection{Prior Selection} \label{Lora_prior}

The choice of an optimal prior is crucial for obtaining a posterior distribution that exhibits optimal performance. Our approach is to assume a normally distributed prior around the pre-trained model, \( p(\theta) = N(\theta; \omega_{\text{pretrained}}, \lambda^{-1}I_{\text{dim}[\theta]}) \), defined with a certain variance \( \lambda^{-1} \). This variance is a hyperparameter that must be carefully chosen based on specific objectives. 

An inappropriate selection of the prior can lead to a posterior that performs poorly on the task at hand. Consequently, we adjust the variance of the prior with the aim of optimizing the posterior quality, as measured by the log likelihood on the validation data. Different choice for the prior variance results in adjusting L2 regularization loss \cite{Hoffmann2021, Balabanov_2023_CVPR}. In the case of LoRA ensembles, this loss is given by
\begin{align} 
\begin{split}
 &\text{L2} = \frac{\lambda}{2} \sum_{i}  ||W^{(i)}-W^{(i)}_\text{pretrained}||^2   = \frac{\lambda}{2} \sum_{i}  ||B^{(i)} A^{(i)}||^2,\\
\end{split}
\label{eq:L2_Lora}
\end{align}
where $A^{(i)}$ and $B^{(i)}$ are LoRA matrices associated with the weights $W^{(i)}$. 

\subsection{Posterior Quality of LoRA Ensembles} \label{Lora_posterior_quality}

LoRA deep ensembles approximate the true posterior over model parameters for the fine-tuning task via variational inference, and the trustworthiness of the resulting uncertainty estimates depends on how closely this approximation matches \( p(\theta | \mathcal{D}_{\text{fine-tune}}) \).
 Hamiltonian Monte Carlo (HMC) remains the gold standard for posterior estimation because it generates nearly exact samples and thus yields the most faithful uncertainty estimates. In principle, one would evaluate any approximation by measuring its agreement and variance relative to the HMC posterior. In practice, HMC-derived posteriors do not always correlate with the best predictive performance: they can be poorly calibrated and sensitive to distributional shifts~\cite{Izmailov2021}, and stochastic ensembles can outperform them in accuracy and NLL~\cite{Balabanov_2023_CVPR}. This does not diminish their status as the best available posterior approximation, and as such they are the target for any potential approximation method. This closeness can be quantified using agreement and variance metrics \cite{Izmailov2021}. Unfortunately, running HMC on large language models and datasets is prohibitively expensive, placing it beyond our computational budget and preventing direct comparison.

Although direct comparison is not possible, prior work has shown that multimodal methods such as deep ensembles yield accurate posterior approximations and significantly outperform single-mode approaches \cite{Fort2019, Izmailov2021, Balabanov_2023_CVPR}. Building on these insights, we argue that our LoRA ensemble–based method delivers high-quality posterior estimates under realistic training and inference constraints, while also surpassing single-mode baselines such as the Laplace approximation, MC dropout, and related techniques. We acknowledge that Yang et al.~\cite{yang2024bayesian} examined LoRA-tuned LLMs primarily through a single-mode Laplace approximation, comparing it against several other methods—including a small ensemble—and demonstrated its performance benefits. Although their results are pertinent to performance-centric applications, they do not address the uncertainty quantification focus of our study. In fact, the Laplace method’s apparent robustness to distribution shifts and “good” calibration may actually indicate a poor approximation of the true posterior~\cite{Izmailov2021}.

\section{Uncertainty quantification}

\subsection{Epistemic model and aleatoric data uncertainty}

Uncertainty quantification is an important topic that assesses the reliability of model predictions \citep{Gal2016}. In general, the total uncertainty in a model’s prediction can be attributed to multiple sources. One widely used categorization is the decomposition into epistemic (or model) uncertainty—arising from insufficient knowledge or data to fully learn the underlying distribution—and aleatoric (or data) uncertainty—stemming from the inherent noise and randomness in the data generation process.

An example of aleatoric data uncertainty is given by the input sequence ``There is a single item in the box. The color of that item is ''. One would expect a model trained on a large corpus of text to produce a predictive distribution with support on the different colors. This posterior distribution tells us that the model is fairly certain that the next token is a color, but that there is an uncertainty about which color. The uncertainty about the color cannot be decreased by improving the training data; it is inherent in the input sequence. 

On the other hand, the input sequence ``Large language models are '' presents a simple conditional completion task, but we would not trust the output predictions from a model trained on a text corpus compiled before the concept was introduced. This epistemic model uncertainty is not captured in the output distribution, but rather in the uncertainty of the model parameters themselves. This model uncertainty can be systematically calculated within a Bayesian formalism, where it corresponds to the shape of the posterior distribution of model parameters conditioned on the training data.

\subsection{Entropic uncertainty metrics}

Depending on the approach used to estimate uncertainty, existing methods can be broadly divided into intrinsic and extrinsic methods. Extrinsic methods involve post-hoc analysis with auxiliary models to estimate uncertainty \cite{lahlou2023deupdirectepistemicuncertainty, kristiadi2021learnableuncertaintylaplaceapproximations}. In contrast, intrinsic methods incorporate uncertainty estimation directly within the model. Here, we employ a common intrinsic methodology based on entropic metrics \cite{malinin_uncertainty_2021} to evaluate uncertainty within a Bayesian formulation, applying it to our LoRA ensemble approximation of the Bayesian posterior for fine-tuned models.

The total predictive uncertainty, encompassing both aleatoric data and epistemic model components, can be quantified using the total predictive entropy:
\begin{align} 
\begin{split}
&H(t^*|s^*, \mathcal{D}) = -\sum_{c} p(t^*=c|s^*, \mathcal{D}) \log p(t^*=c|s^*, \mathcal{D}),\\
\end{split}
\label{eq:entropy}
\end{align}
where \(s^*\) is the input string, \(\mathcal{D}\) is the training dataset, and \(c\) denotes all possible output tokens. The output probabilities \(p(c \mid s^*, \mathcal{D})\) are computed by averaging softmax outputs across model parameters \(\theta\) sampled from the Bayesian posterior \(p(\theta \mid \mathcal{D})\).

Epistemic model uncertainty is defined as the variability in predictions that arises from sampling model parameters \(\theta\) from the posterior distribution \(p(\theta \mid \mathcal{D})\). This variability captures how the distribution of predicted outputs shifts with each parameter set, weighted by its posterior probability. A common metric for this purpose is the average mutual information between the model parameters and output, conditioned on the training data \cite{MacKay1992, malinin_uncertainty_2021}:
\begin{align} 
\begin{split}
&\text{MI}(\theta, t^*|s^*, \mathcal{D}) \\
&= H(t^*|s^*, \mathcal{D}) - \mathbb{E}_{\theta\sim p(\theta|\mathcal{D})} \left[ H(t^*|s^*, \theta) \right].\\
\end{split}
\label{eq:MI}
\end{align}
For ensemble posteriors, zero mutual information implies that all ensemble members yield the same predictive distribution, indicating no variability in the outputs and thus zero epistemic uncertainty. On the contrary, if the mutual information is high, there is significant variability in the predictive distribution across the ensemble, reflecting high epistemic uncertainty.

Using these metrics, we can identify three regimes that describe how the model perceives data:

\begin{itemize}
    \item \textit{Known-certain}: Low total and epistemic uncertainty; the posterior ensemble strongly agrees on a narrow, confident prediction distribution.
    \item \textit{Known-uncertain}: High total uncertainty, low epistemic uncertainty; uncertainty arises from data noise or model expressiveness limitations, with the ensemble members agreeing on a broad distribution.
    \item \textit{Unknown}: High total and epistemic uncertainty; the posterior members disagree on the prediction distribution, highlighting a lack of knowledge.
\end{itemize}

In this work, we use uncertainty quantification from entropic metrics of the LoRA ensemble posterior approximation to better understand how fine-tuning reshapes the knowledge of the model, revealing shifts in data characteristics and how the model adapts to new information. In particular, we study the dynamics of the uncertainty metrics during fine-tuning.

\section{Numerical Results} 
\subsection{Datasets} 

We evaluate our methods on three multiple-choice QA datasets: CommonsenseQA (CQA) \cite{talmor-etal-2019-commonsenseqa}, and the Social Sciences (SS) as well as STEM components from MMLU \cite{hendryckstest2021}. The ensemble is trained on the CQA training dataset but tested on three different validation datasets: the (in-domain) CQA validation dataset and the (out-of-domain) MMLU STEM and MMLU SS validation datasets. Detailed information about the training and validation (test) splits is available in Appendix \ref{ap_datasets}.

\subsection{LoRA Ensembles} 

We train LoRA ensembles using the pre-trained 7-billion-parameter Mistral-7B model~\cite{jiang2023mistral}. The LoRA parametrization is implemented using PEFT~\cite{peft}. We evaluate ensembles with sizes \(M \in \{1, 5, 10, 20\}\) in our experiments and ablation studies. Following the method of Hu et al.~\cite{hu2021lora}, we apply the adapter \(\Delta W = \alpha BA\) exclusively to the query and value matrices within the self-attention modules of Mistral-7B, with \(\alpha = 32\). The adapter rank is set to \(r = 8\), resulting in \(3{,}407{,}872\) trainable parameters. The adapter matrices \(B\) are initialized to zero, and the entries of \(A\) are initialized using a Kaiming-uniform distribution~\cite{he2015delving}.

The ensemble members are trained using the prescribed regularisation from Section~\ref{Lora_prior}, corresponding to using the pretrained Mistral-7b model as a prior. We use the Adam optimizer, training for 10 epochs with learning step sizes of \(5\cdot10^{-6}\) and batch sizes of 8. This learning rate is smaller compared to those in \cite{wang2023lora} to better refine the learning curve before the models become overfitted. The token size limit is fixed at 128 for CQA and 512 for MMLU. We performed an ablation study on L2 regularization \(\lambda/2 = 0.01, 0.1, 1, 10\), finding optimal performance at \(\lambda/2 = 1\). The training focuses solely on the single token output representing the answer, with all other token outputs being masked. 

To calculate performance metrics and uncertainty estimates, we reduced the output dimension from 32,000 to 6. Five of these dimensions represent QA choice tokens (a, b, c, d, e), and the sixth aggregates the softmax prediction scores associated with all other tokens. We found that, even before fine-tuning, the pretrained Mistral-7b checkpoint assigns a near-zero probability to the sixths class with very few exceptions. This suggests that the model reliably comprehends the format of multiple-choice questions and answers, consistently producing appropriate tokens.

\subsection{Performance metrics}

\begin{figure}[t]
\begin{center}
\centerline{\includegraphics[width=0.7\columnwidth]{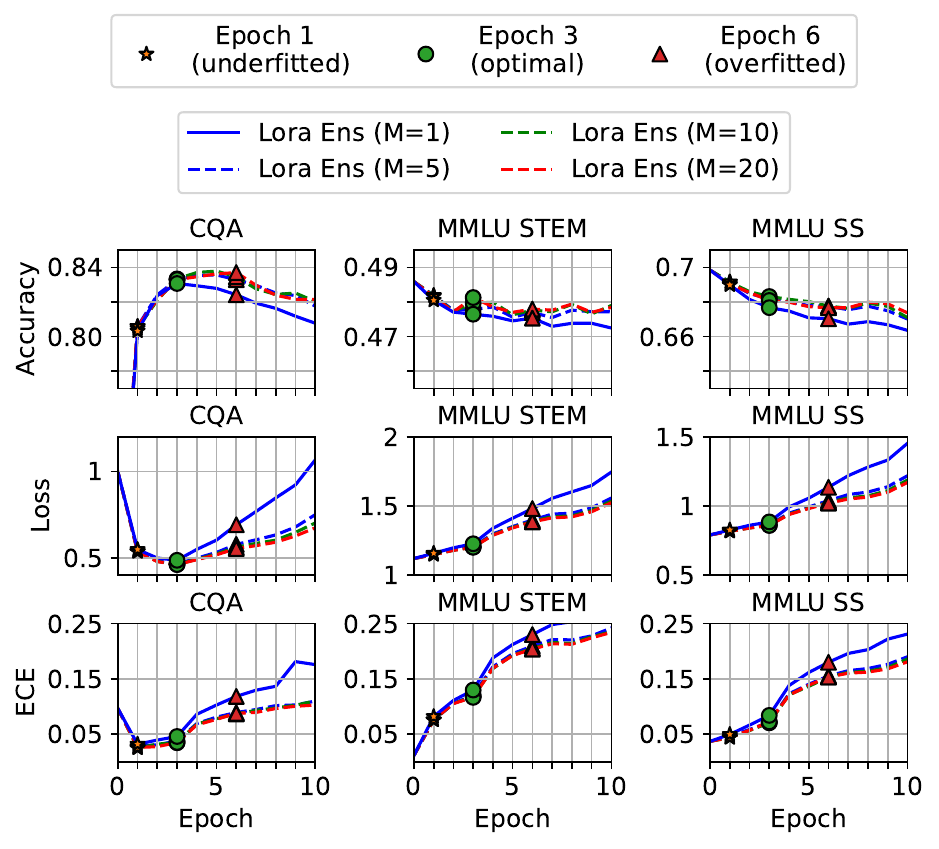}}
\caption{Performance of LoRA ensembles trained on CQA and evaluated on the CQA, MMLU STEM, and MMLU SS datasets. Metrics (computed over the ensemble mean distribution) include accuracy, negative log-likelihood (NLL) loss, and expected calibration error (ECE). We show ensemble sizes \(M \in \{1,5,10,20\}\); for \(M=1,5,10\), results are averaged over 20, 4, and 2 independent realizations, respectively, while \(M=20\) uses a single run. Metric values at epoch 1 (underfitting), epoch 3 (optimal), and epoch 6 (overfitting) are highlighted.}
\label{icml-metrics}
\end{center}
\end{figure}

In Fig.~\ref{icml-metrics}, we present the performance metrics for LoRA ensembles with ensemble sizes \(M = 1\), \(5\), \(10\), and \(20\). The ensembles are trained on multiple-choice questions from the CQA training dataset and evaluated on three validation datasets: CQA (first column), MMLU STEM (second column), and MMLU Social Studies (third column).

As shown in Fig.~\ref{icml-metrics}, the loss curves for ensemble sizes \(M = 5\), \(10\), and \(20\) converge closely, indicating that small ensembles (\(M = 5\)) provide posterior approximations comparable to those of larger ensembles (\(M = 20\)). This convergence is particularly interesting and practically relevant, as it suggests that LoRA ensembles can achieve accurate uncertainty estimates without significantly increasing inference costs. Based on this observation, we use \(M = 5\) ensembles for uncertainty quantification in the following section.

The ensemble members start over-fitting after a couple of epochs as can be seen by the validation loss in the second row of panels, but interestingly without considerable drop in accuracy in the first row of panels. As will be seen in more detail in Section \ref{histogram_section}, this signals that the ensemble members become more overconfident on predictions that are wrong, corroborated by the increasing expected calibration error with epoch seen in the bottom row of Fig.~\ref{icml-metrics}.  This is consistent with the recent reports that fine-tuned LLMs often exhibit overconfidence \cite{jiang-etal-2021-know, 2022arXiv220514334L, xiao2022uncertainty, he2023preserving, tian2023just, openai2023gpt4}.

For fine-tuning on CQA, shown in the first column, overfitting is significantly reduced for ensembles compared to individual models, as evidenced by the gap between the loss curves for \(M = 1\) and those for \(M = 5, 10, 20\) at later epochs. This suggests that ensemble members produce more confident yet diverse predictions. This means that ensemble members output more confident, but different predictions. This is a common feature of Bayesian posteriors \cite{Blundell2015, zhang2020cyclical, kristiadi2020bayesian, pmlr-v139-ober21a, fortuin2022bayesian, pmlr-v139-aitchison21a, yang2024bayesian}. This signals high epistemic uncertainty in this regime, indicating that the validation set is perceived as ``unknown" to the model.

\subsection{Dynamics of uncertainty metrics} \label{histogram_section}

We analyze data-resolved dynamics of uncertainty measures using two-dimensional histogram plots of mutual information and entropy \cite{linander2023looking}. This approach allows us to effectively disentangle different sources of uncertainty, providing a detailed view of how the model perceives data across the ``known-certain," ``known-uncertain," and ``unknown" regimes during fine-tuning. By visualizing the relationship between mutual information and entropy, we can track how the model's knowledge evolves, identify patterns in data complexity, and assess the model's ability to adapt to fed information. This method not only highlights the dynamics of learning, but also helps identify limitations of the model, offering actionable insights for further improvement.

\begin{figure}[t]
\vskip 0.2in
\begin{center}
\centerline{\includegraphics[width=0.7\columnwidth]{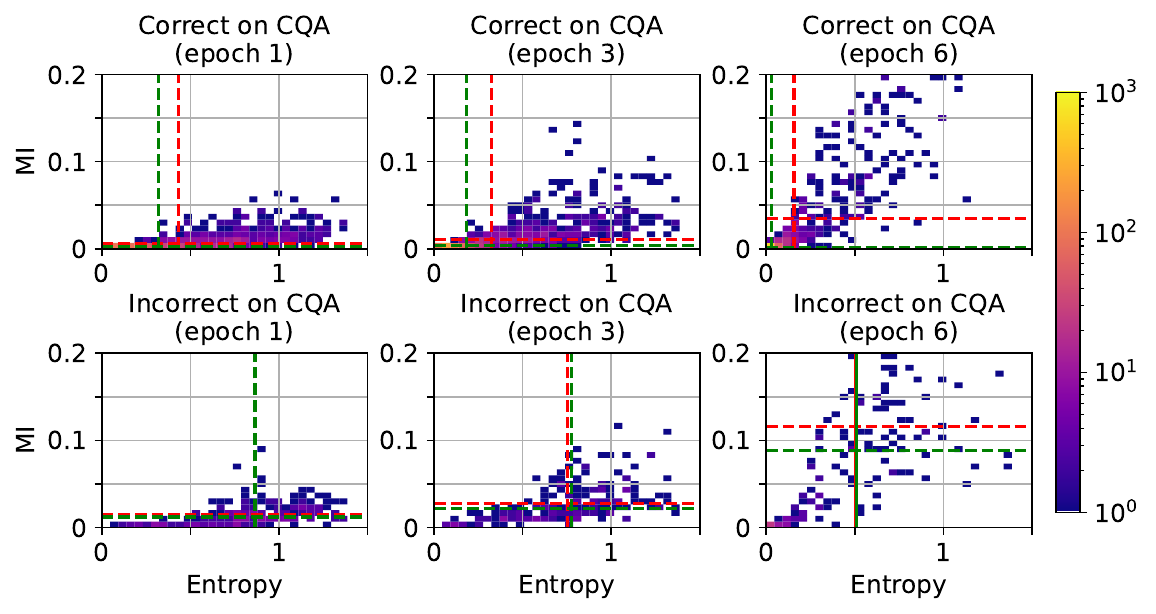}}
\caption{In-distribution evolution of uncertainty during fine-tuning. Histograms of predictive entropy and mutual information for a LoRA ensemble trained and evaluated on the CQA dataset. The ensemble consists of $M=5$ members, with a LoRA rank of 8, $\alpha=32$, and an L2 LoRA loss of~1. This figure illustrates the evolution of uncertainty measures for in-domain data across training epochs (columns), differentiated by correct (top row) and incorrect (bottom row) predictions. We also depict the corresponding mean (red) and median (green) entropy and mutual information values. Color represents the number count in each bin.}
\label{icml-cqa_uncertainty}
\end{center}
\vskip -0.2in
\end{figure}

In Fig. \ref{icml-cqa_uncertainty}, we show the epoch evolution of uncertainty measures for a LoRA ensemble that was fine-tuned and evaluated on the CQA dataset. The mutual information (MI) on the vertical axis, and the predictive entropy (Entropy) on the horisontal axis, are binned over every sample of the validation set of CQA and the number of samples in each bin is indicated by the colorbar. We distinguish between incorrect and correct predictions to enhance interpretability. The ensemble attains optimal validation loss at epoch 3, as shown in Fig.~\ref{icml-metrics}. Epochs 1 and 6 exhibit underfitting and overfitting, respectively. The true posterior \(p(\theta|D)\) should yield nearly zero mutual information when evaluated on the validation dataset, which should belong to ``known" data by design. However, in each of the underfitted, optimal, and overfitted regimes, there is a fraction of validation samples that exhibit a moderate spread in mutual information. Samples with high mutual information indicate that parts of the target domain are perceived as “unknown” by the ensemble, highlighting regions where expanding or refining the training data could most improve performance.  

By examining the mean and median of the entropic metrics in Fig. \ref{icml-cqa_uncertainty}, we can observe overall patterns in the validation data perception by the ensemble. Interestingly, the ensemble becomes progressively more confident on correct QAs, reaching its highest confidence once the ensemble overfits, where most correct QAs are predicted with almost zero uncertainty. Although overall performance (as measured by loss shown in Fig.~\ref{icml-metrics}) degrades at epoch 6 compared to epoch~3, the model has actually learned the bulk of the validation dataset more effectively. The loss increase arises because the ensemble has also become more confident about incorrect QAs, increasing the cross-entropy. However, almost all of these incorrect predictions are marked by high mutual information, placing them in the “unknown” regime. Since these points can be identified by their high mutual information, they can be filtered out at inference time or used to refine the training procedure (e.g., by including more instances of these data). This result is noteworthy because it shows that the ensemble continues to improve even after reaching its minimum loss at epoch 3. By epoch 6, it performs better on most of the dataset, although its increased confidence in the remaining poorly performing data inflates the overall loss. Nonetheless, these problematic data points remain mostly distinguished by high epistemic uncertainty.

\begin{figure}[t]
\vskip 0.2in
\begin{center}
\centerline{\includegraphics[width=0.7\columnwidth]{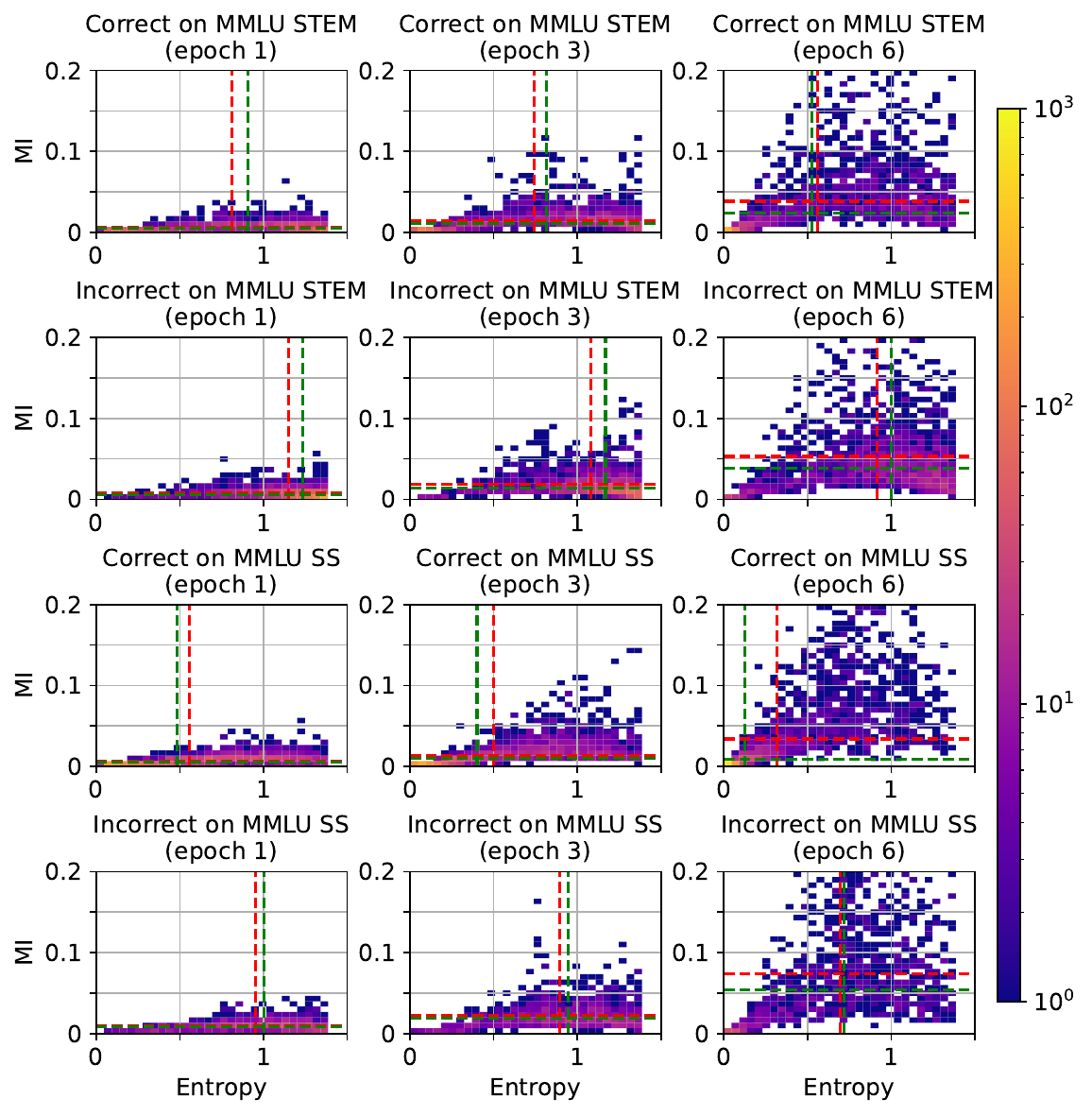}}
\caption{Out-of-distribution evolution of uncertainty during fine-tuning. Histograms of predictive entropy (Entropy) and mutual information (MI) for a LoRA ensemble trained on CQA and evaluated on MMLU STEM (top rows) and MMLU SS (bottom rows) datasets. The ensemble consists of $M=5$ members, with a LoRA rank of 8, $\alpha=32$, and an L2 LoRA loss of~1. These panels illustrates the evolution  of uncertainty measures for out-of-training-domain data across training epochs (columns left to right), differentiated by correct (first and third row) and incorrect (second and forth row) predictions. We also depict the corresponding mean (red) and median (green) entropy and mutual information values. Color represents the number count in each bin.}
\label{icml-mmlu_uncertainty}
\end{center}
\vskip -0.2in
\end{figure}

Fig.~\ref{icml-mmlu_uncertainty} shows the uncertainty histograms for MMLU STEM and MMLU SS, calculated using an ensemble trained on CQA. Starting with MMLU STEM in the first and second row of panels, we draw two key insights. First, among the wrongly classified samples for the optimal epoch 3 (second row, second column), a large fraction falls into the ``known-uncertain`` regime, characterized by high predictive entropy but low mutual information. This is in contrast to MMLU SS, which exhibits a more moderate spread of predictive entropy in the corresponding panel (last row, second column). This observation highlights the relatively higher complexity of STEM compared to SS datasets. The ensemble perceives the STEM samples as equally ``known" as SS samples (similar mutual information) but assigns them broader predictive distributions, indicating greater ``uncertain" characteristics due to a lack of precise knowledge. 
\begin{figure}[t]
\vskip 0.2in
\begin{center}
\centerline{\includegraphics[width=0.7\columnwidth]{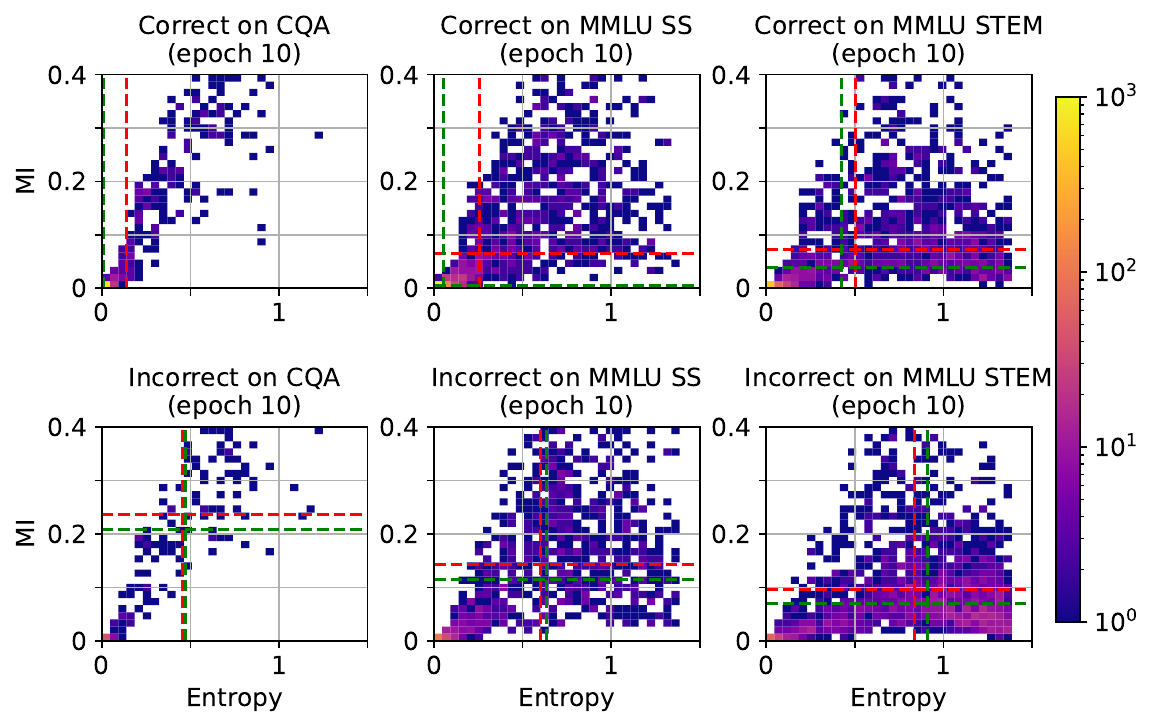}}
\caption{Late overfitting regime (epoch 10). Histograms of predictive entropy (Entropy) and mutual information (MI) for a LoRA ensemble trained on CQA (left column) and evaluated on MMLU SS (middle column) and MMLU STEM (right column) datasets. The ensemble consists of $M=5$ members, with a LoRA rank of 8, $\alpha=32$, and an L2 LoRA loss of~1. There is still a significant fraction of samples (lower right panel) in MMLU STEM with low mutual information and high total uncertainty.
}
\label{icml-mmlu_uncertainty_ep10}
\end{center}
\vskip -0.2in
\end{figure}

Second, continuing with MMLU SS in the third and fourth rows of Fig.~\ref{icml-mmlu_uncertainty}, we observe a significant decrease in predictive entropy as the number of epochs increases, indicating that the ensemble learns features that enable it to classify samples with greater confidence. This behavior is similar to CQA in Fig.~\ref{icml-cqa_uncertainty}, but the key difference is that, in this case, the data is from an out-of-domain dataset. This demonstrates that the ensemble continues to improve on the majority of MMLU SS samples, despite not being specifically trained on them, suggesting a similarity between the CQA and MMLU SS domains.

Interestingly, the evolution of entropic quantities reveals a key difference between the datasets. For MMLU STEM, there is a notable density of ``known-uncertain" samples (high entropy, low mutual information) (see Fig.~\ref{icml-mmlu_uncertainty}, second row), whereas for MMLU SS, high-entropy samples are more evenly distributed in terms of mutual information. The density of ``known-uncertain" MMLU STEM samples largely persists despite overfitting (see Fig.~\ref{icml-mmlu_uncertainty_ep10}), suggesting that the fine-tuned model ensemble perceives confusing samples from MMLU STEM as being more consistently ``known-uncertain," whereas confusing samples from MMLU SS are forgotten more rapidly, transitioning into the ``unknown" regime.

\section{Conclusions} 
We derived a principled, and practically efficient method for posterior approximation of fine-tuned LLMs using ensembles of low-rank adaptation models. Using the posterior approximation, we derived uncertainty quantification using entropic metrics, and followed their evolution during fine-tuning.
This enabled us to draw a number of conclusions regarding the balance of retained knowledge and domain specific adaptation. First, the low-rank ensemble shows retained accuracy while also decreasing both predictive uncertainty and mutual information in the overfitting regime. At the same time, incorrect predictions are increasingly correlated with higher mutual information, opening up for inference time filtering and active learning.
Second, during fine-tuning towards CQA, there is a distinct difference between the change in uncertainty for MMLU STEM and MMLU social science. For the STEM dataset, there is a large fraction of samples with low mutual information and high predictive entropy, hinting at in-domain but uncertain predictions. This can be connected to the lower prior accuracy of MMLU STEM compared to MMLU SS and CQA, indicating low representative power of the model in this domain. These STEM samples exhibit increased robustness, through broad yet consistent predictive distributions across ensemble members, to a large extent surviving overfitting.

As fine-tuning continues to be an important tool in shaping model output \cite{zhang2023instruction}, it is also increasingly important to understand how the knowledge of the pre-trained model changes as the model is adapted to a target domain. The methods presented here are generally applicable to different fine-tuning scenarios, and provide a systematic treatment of prediction uncertainty for fine-tuned LLMs.

\section{Acknowledgments}
The computations were enabled by resources provided by the National Academic Infrastructure for Supercomputing in Sweden (NAISS) and
the Swedish National Infrastructure for Computing (SNIC)
at C3SE partially funded by the Swedish Research Council
through grant agreements 2022-06725 and 2018-05973.
\section{Impact Statement}

This paper advances the field of Machine Learning by enhancing the reliability and understanding of large language models (LLMs) through uncertainty quantification using LoRA ensembles. Our work contributes to the understanding and trustworthiness of LLM predictions, a step forward in AI reliability and applicability.

\bibliography{refs}
\bibliographystyle{iclr2025_conference}

\newpage
\appendix

\onecolumn
\section{Datasets} \label{ap_datasets}

\subsection{Size}
For MMLU STEM and Social Studies datasets, we use the original test split as the validation set, as in \cite{wang2023lora}. For CQA dataset we use the standard training and validation splits. The topics in the MMLU dataset are categorized into STEM, Social Studies, Humanities, and Other, following the classification defined in \cite{hendryckstest2021}. 
\begin{table}[h]
\centering
\begin{tabular}{l|l|l}
\hline
\textbf{Dataset} & \textbf{Train Size} & \textbf{Validation Size} \\ \hline
CQA & 9741 & 1221 \\ \hline
MMLU SS   & 397  & 3077 \\ \hline
MMLU STEM & 430  & 3153 \\ \hline
\end{tabular}
\caption{Summary of the CQA and MMLU dataset sizes.}
\label{dataset_sizes}
\end{table}
\subsection{Examples}
The datasets have been preprocessed into a single, consistently formatted prompt question-answer. Below, we present examples of these questions for each of the considered dataset. 

 \begin{mdframed}[backgroundcolor=gray!20, linecolor=black, linewidth=1pt, roundcorner=10pt]
\large

Q: A revolving door is convenient for two direction travel, but it also serves as a security measure at a what?

Answer Choices:

(a) bank

(b) library

(c) department store

(d) mall

(e) new york

A: (a).

\smallskip
\hfill CommonsenseQA
\end{mdframed}

\begin{mdframed}[backgroundcolor=gray!20, linecolor=black, linewidth=1pt, roundcorner=10pt]
\large

Q: Which one of the following is the most appropriate definition of a 99\% confidence interval?

Answer Choices:

(a) 99\% of the time in repeated samples, the interval would contain the true value of the parameter

(b) 99\% of the time in repeated samples, the interval would contain the estimated value of the parameter

(c) 99\% of the time in repeated samples, the null hypothesis will be rejected

(d) 99\% of the time in repeated samples, the null hypothesis will not be rejected when it was false

A: (a).

\smallskip
\hfill MMLU SS
\end{mdframed}

\begin{mdframed}[backgroundcolor=gray!20, linecolor=black, linewidth=1pt, roundcorner=10pt]
\large

Q: Find all c in Z\_3 such that Z\_3[x]/(x\textasciicircum2 + c) is a field.

Answer Choices:

(a) 0

(b) 1

(c) 2

(d) 3

A: (b).

\smallskip
\hfill MMLU STEM
\end{mdframed}

\end{document}